\documentclass[conference]{IEEEtran}
\IEEEoverridecommandlockouts
\usepackage{cite}
\usepackage{amsmath,amssymb,amsfonts}
\usepackage{algorithmic}
\usepackage{graphicx}
\usepackage{textcomp}
\usepackage{xcolor}
\usepackage{mathtools}
\usepackage{commath}
\usepackage[hyphens]{url}
\usepackage{subcaption}
\usepackage{balance}

\begin{document}
\bstctlcite{IEEEexample:BSTcontrol}

\title{Analyzing Scientific Publications using Domain-Specific Word Embedding and Topic Modelling}

\author{
\IEEEauthorblockN{Trisha Singhal}
\IEEEauthorblockA{\textit{SUTD-MIT International Design Centre} \\
\textit{Singapore University of Technology and Design}\\
Singapore \\
trisha\_singhal@sutd.edu.sg}
\and
\IEEEauthorblockN{Junhua Liu}
\IEEEauthorblockA{\textit{Information Systems Technology and Design Pillar} \\
\textit{Singapore University of Technology and Design}\\
Singapore \\
junhua\_liu@mymail.sutd.edu.sg}
\and

\IEEEauthorblockN{Lucienne T.M. Blessing}
\IEEEauthorblockA{\textit{Engineering Product Development Pillar} \\
\textit{Singapore University of Technology and Design}\\
Singapore \\
lucienne\_blessing@sutd.edu.sg}
\and
\IEEEauthorblockN{Kwan Hui Lim}
\IEEEauthorblockA{\textit{Information Systems Technology and Design Pillar} \\
\textit{Singapore University of Technology and Design}\\
Singapore \\
kwanhui\_lim@sutd.edu.sg}
}

\IEEEoverridecommandlockouts
\IEEEpubid{\makebox[\columnwidth]{978-1-6654-3902-2/21/\$31.00~\copyright~2021 IEEE \hfill} \hspace{\columnsep}\makebox[\columnwidth]{ }}

\maketitle

\IEEEpubidadjcol

\begin{abstract}
The scientific world is changing at a rapid pace, with new technology being developed and new trends being set at an increasing frequency. This paper presents a framework for conducting scientific analyses of academic publications, which is crucial to monitor research trends and identify potential innovations. This framework adopts and combines various techniques of Natural Language Processing, such as word embedding and topic modelling. Word embedding is used to capture semantic meanings of domain-specific words. We propose two novel scientific publication embedding, i.e., PUB-G and PUB-W, which are capable of learning semantic meanings of general as well as domain-specific words in various research fields. Thereafter, topic modelling is used to identify clusters of research topics within these larger research fields. We curated a publication dataset consisting of two conferences and two journals from 1995 to 2020 from two research domains. Experimental results show that our PUB-G and PUB-W embeddings are superior in comparison to other baseline embeddings by a margin of $\sim$0.18-1.03 based on topic coherence.
\end{abstract}


\section{Introduction}
The scientific and technological worlds are changing at an unprecedented rate, thus increasing the importance of monitoring research trends to identify innovation potential. Trend research and identification can be done using a variety of sources, with scientific literature books, articles, and publications receiving substantial attention from researchers worldwide~\cite{guenther2017science,preiser2019identifying,mohammadi2020exploring}. The analysis of publications has proved useful in identifying emerging topics and tracking their growth or decline over the years using linguistic features. The use of Natural language Processing (NLP) techniques to support this analysis makes it easier to discover  patterns and allow for answering more specific research questions. 

Word Embedding is an important and widely used NLP technique to identify the semantic meanings of a text corpus. These semantic meanings are useful for identifying and quantifying the word-word similarities and global contextual meaning of text corpora.
An increasing number of word embeddings can be found in the literature, such as count vectorizers \cite{count_vect} and TF-IDF \cite{tfidf_vect}, which are more classical word representation techniques. These classical approaches are linear language modeling approaches and often fail to model the true contextual meaning of text corpora. In contrast, Word2Vec \cite{mikolov2013efficient}, GloVe \cite{pennington2014glove}, and ELMO \cite{peters2018deep} are some of the more modern techniques of contextualizing meanings of text corpora, which incorporate neural networks for non-linear language modelling. However, these models are often trained on datasets derived from Twitter, Wikipedia, or general pieces of text and are therefore not entirely suitable for the analysis of scientific publications due to the existence of domain-specific words in these corpora. With this motivation, we present two domain-specific word embeddings termed PUB-G and PUB-W. We use these embeddings to cluster scientific publications based on their abstract to identify various areas of research in the respective domain. We then use topic modeling to identify more detailed research topics in these areas. 

The paper makes the following contributions: 

\begin{enumerate}
    \item We curate a publication dataset for two conference series and two journals of two different disciplines with a total of 10.4k publications for the period 1995 to 2020.
    \item We propose novel domain-specific embeddings based on GloVe and, alternatively, explore an embedding based on Word2Vec for publication data termed as PUB-G and PUB-W respectively. We further used these embeddings to cluster publications based on their abstracts. 
    \item We develop a baseline classical approach for publication clustering and compare this approach with other competitive baseline embeddings.
    \item We show that the research topics identified by PUB-G embedding show a better coherence score. 
\end{enumerate}

The remainder of this paper is organized as follows. Section~\ref{relatedWork} discusses the related works. Section~\ref{framework} shows the workflow of our proposed framework and provides a detailed explanation of each component, and Section~\ref{dataset} describes our dataset. Section~\ref{results} discusses the experimental results and main findings. The conclusions can be found in Section~\ref{conclusion}.

\section{Related Work}
\label{relatedWork}
NLP has widespread use and is being applied in a myriad of tasks ranging from language translation~\cite{bahdanau2014neural,takahashi2020automatic,arivazhagan2020re}, sentiment analysis of text~\cite{bakshi2016opinion,ke2020sentilare,basiri2021abcdm}, document analysis~\cite{bowen2009document,mendsaikhan2019identification}, and social media analytics~\cite{liu2020epic, liu2020crisisbert,George-JBigData21}. In the following sections, we discuss some related works in several relevant sub-fields of NLP.

\subsection{Classical Document Analysis}
Digital document analysis has been a research field for several years. \cite{mao2003document} provides a detailed discussion of the traditional approaches that were used to analyze the structure of electronic documents. On the other hand, there are various tools available in the market today to perform information extraction from scientific literature \cite{councill2008parscit, lopez2009grobid, tkaczyk2014cermine, singh2016ocr++}. Many researchers used traditional approaches like Support Vector Machines (SVM), Latent Dirichlet Allocation (LDA), Singular Value Decomposition (SVD) and Hidden Markov Model (HMM) to implement various text analysis techniques. \cite{takasu2003bibliographic} used an extended HMM to extract the bibliographic attributes from the references, \cite{han2003automatic} used SVM classifier for two-stage metadata extraction from headers of research publications, and \cite{chiarello2018automatic} applied an ensemble ML approach to automatically extract users from patents.

\subsection{Deep Learning-based Document Analysis}
The application of neural networks (NN) to digital documents helped enormously in extracting and analyzing the documents and gain valuable insights. On this account, \cite{stahl2018deeppdf} extracted text information by identifying various sections of scientific publications in the form of PDF documents using deep learning-based NN, U-Net. Some researchers worked beyond the textual information such as \cite{yang2017learning}, who used end-to-end multimodal fully convolutional neural networks to perform pixel-wise page segmentation to extract semantic features of the document. Other researchers~\cite{clark2015looking, clark2016pdffigures} extracted
figures in research papers at NIPS, ICML and AAAI.

\subsection{Trend Analysis using Publications}
Trend analysis has been a significant research topic in several fields. Recently, Ordun et. al. \cite{ordun2020exploratory} did a thorough analysis of COVID-19 tweets using topic modeling and pattern matching to identify high-level trends, events with sudden spikes, distinctive topics, speed of tweeting and re-tweeting, and network behaviors. On a similar topic, Kwan and Lim~\cite{Kwan-IUI21,Kwan-ASONAM20} used sentiment analysis, topic modeling and temporal analysis techniques on tweets to study trends and discussions about COVID-19 in various countries. Schoch et. al. \cite{schoch2017topic} explored literary genre using topic modeling. Chiarello et. al.\cite{chiarello2019text} using state-of-the-art text mining techniques to analyze research papers published in the Engineering Design field to identify the evolution of various research themes. Pek and Lim used academic publications to identify key business trends, particularly the various popular topics and the frequency of these topics over the years~\cite{Pek-BigData19}. Similarly, research publications were used in HCI: Yang et. al. \cite{yang2018mapping} visualized the use of ML to improve user experience (UX), Carter et. al. \cite{carter2014paradigms} analyzed the understanding of games and play research within four research paradigms in the field of HCI, and \cite{valdez2019trends} studied the emerging trends and changes in HCI over a decade. 

\subsection{Textual Word Embeddings}
Textual word embeddings are often used in NLP research for language modeling~\cite{li2018word,almeida2019word}. These embeddings transform textual words into an n-dimensional vector space, which are useful to 1) quantify word-word similarity and 2) model global contextual meanings of text corpora. There are various techniques for developing such a language model. Some of the classical methods in the literature are count vectorization and TF-IDF vectorization. These methods are based on word frequencies. Word2Vec presents a shallow neural network-based approach that optimizes to predict a masked word given the words before and after the masked word~\cite{mikolov2013efficient}. GloVe presents a regression-based model to predict the conditional probability of a word appearing given another word. Context-aware word embeddings, such as such as Embeddings from Language Model (ELMo)~\cite{peters2018deep} and Bidirectional Encoder Representations from Transformers (BERT)~\cite{devlin2018bert}, were more recently proposed to generate word representations that better consider the context of the sentence. However, all these embeddings are usually trained on common text corpora~\cite{pennington2014glove}. Thus, these existing embeddings are not suitable for analyzing scientific publications which often are abundant with domain-specific words. 

\section{Proposed Framework}
\label{framework}
\begin{figure*}
    \centering
    \includegraphics[width=\linewidth, trim=15mm 0mm 5mm 0mm]{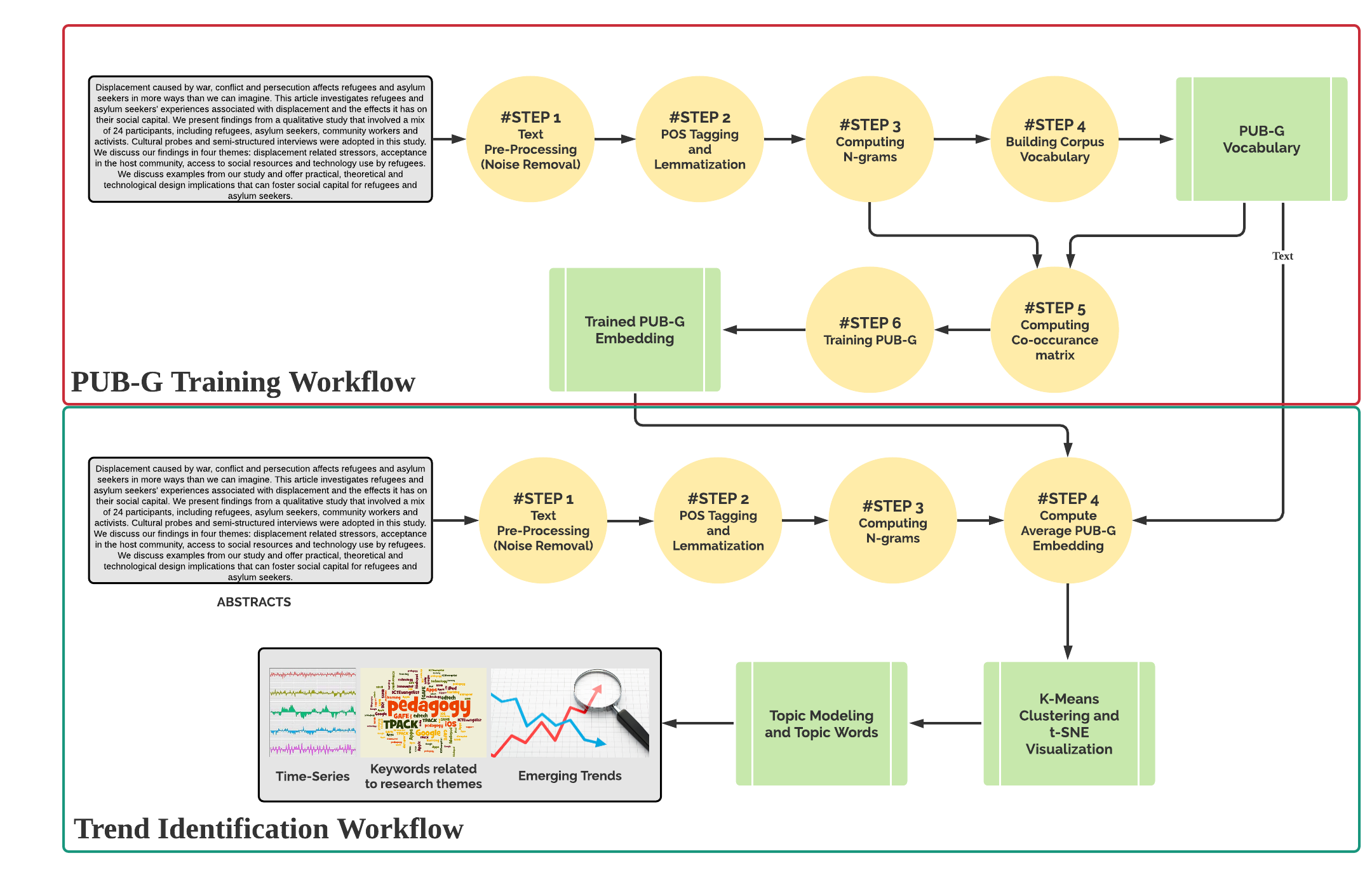}
    \caption{Framework for analyzing academic publications using word embedding and topic models.}
    \label{fig:flowchart}
\end{figure*}

Our analysis is based on the abstracts of publications from two conferences and two journals in the Human-Computer Interaction (HCI) and Engineering Design research fields (more details later). We use a classical and neural network-based hybrid linguistic analysis approach to uncover trends in these conferences and journals. Figure \ref{fig:flowchart} shows our analysis methodology which will be explained in detail in this section.

\subsection{Topic Clustering with Abstracts}

\subsubsection{\textbf{Text Pre-Processing}}
The abstracts are pre-processed by reducing noise to facilitate the subsequent analyses. The pre-processing consists of the following steps:

\textbf{Hyperlinks.} The first step performed to clean the dataset is the removal of hyperlinks. Regex patterns are implemented for the same. These consist of a small sequence of characters defining a specified syntax that is used to match all the possible sets of strings in a given text. These regular expressions are supported and accessed by the Python module \textit{'re'}. 

\textbf{Punctuations.} Next, all the punctuation symbols including `[', `,', `\textbackslash', `.', `!', `?', `]' are replaced by an empty string using regular expressions (regex).

\textbf{Numeric Values.} Numerical values such as dates, amounts, etc. do not contribute much information for our purpose and hence, we removed numbers from the documents, again using Regex. 

\textbf{Lowercase.} To prevent the model from being case-sensitive, we converted the text in lowercase using string method, \textit{lower()}.

\textbf{Whitespaces.} It is essential to remove unnecessary whitespaces from the data to reduce noise. This is done by using the string method, \textit{strip()} that removes leading and trailing whitespaces.

\subsubsection{\textbf{Tokenization}}

The tokenization process assigns a unique identifier to each unique word in the publication corpus. This is usually done as a preliminary step in many of the natural language processing pipelines for obtaining language features. We used \emph{Gensim} library to perform tokenization on the corpus.

\subsubsection{\textbf{Part-Of-Speech Tagging and Lemmatization}}
Part-Of-Speech (POS) tagging is used to allocate each token a POS tag, such as noun, adjective, verb, and adverb based on its contextual interpretation. 
Following which, lemmatization transforms all tokens from their grammatical modulation to root form.

\subsubsection{\textbf{N-grams}}
In computational linguistics, an n-gram is identified as a continuous pattern of \textbf{n} words \textit{}in a text corpus. An n-gram of size 1 is referred to as unigram, size 2 is referred to as bi-gram, and size 3 as trigram. Identifying such patterns in texts are often necessary to effectively uncover contextual meanings of language . Some of the most common bigrams and trigrams identified in our datasets are \textit{`augmented\_reality'}, \textit{`privacy\_concern'}, \textit{`computer\_mediated\_communication'}, etc. It is evident that  if these words are extracted as  unigrams, their linguistic meaning is lost. 

\subsection{Text Embeddings}
\subsubsection{\textbf{Baseline Textual Embedding}}
Text embedding is an $N$ dimensional vector for each unique word in the corpus. In other words, it stipulates words into $N$ dimensional vector space, from which, e.g.  the semantic similarity among different words can be derived. We use Term Frequency-Inverse Document Frequency (TF-IDF)~\cite{salton1988term} as our baseline textual embedding which essentially is a 1-dimensional text embedding. TF-IDF score for the word $t$ in document $d$ from the document set D is obtained as follows:

\begin{equation}\label{eq:tfidf}
  \begin{gathered}
  TFIDF (t,d,D) = tf(t,d)\cdot idf(t,D) \\
  where,\;\; tf (t,d) = log(1+freq(t,d)) \\
  and,\;\; idf (t,D) = log\frac{N}{count(d \in D: t \in d)}
  \end{gathered}
\end{equation}

Let us assume that there are $T$ unique words, bigrams or trigrams in the publication corpus. Let $t_i$ be the $i$\textsuperscript{th} unique word. Our textual embedding is $T$ dimensional where $s_i$ is the $i$\textsuperscript{th} value of the $T$ dimensional vector. $s_i$ for the abstract $d$ is calculated as follows: 

\begin{equation}\label{eq:embedding}
  \begin{gathered}
    s_i = TFIDF(w_i,d,D) \\
    S = [s_i]\;\; \forall i \in [1:T]
  \end{gathered}
\end{equation}

$S$ is used as a baseline for textual embedding. Our publication corpus has $15,125$ unique TF-IDF features for ICED, $21,066$ for CHI, $6,026$ for TOCHI, and $4,998$ for RIED. 

\subsubsection{\textbf{PUB-W Embedding}}
We explored another textual embedding, Continuous Bag-of-Words (CBOW), Word2Vec \cite{mikolov2013efficient}. The model contains a two-layer neural network for training that includes an input, a hidden, and an output layer. The input is given in a vector form developed by converting words into vectors using one-hot encoding. The hidden layer is a dense (fully-connected) layer with word embeddings as the weights and the output layer uses a softmax classifier to generate the probabilities for the target words. 

\begin{equation}
\begin{split}
    minimize~J_n = -\frac{1}{T}\sum _{t=1} ^ {T} {log P(w_t|w_{t-n},...,w_{t-1},}\\ {w_{t+1},...,w_{t+n})}
\label{eq:word2vec}
\end{split}
\end{equation}

Assuming the input text has $T$ words. For each $t \in [1:T]$, optimization function first computes the log of conditional probability of predicting $t$\textsuperscript{th} word given the previous $n$ number of words and $n$ number of words after the $t$\textsuperscript{th} word. Then, it computes the sum of the log of conditional probabilities for each word in the text. Finally, the objective is to minimize the negative of this summation which is represented in equation \ref{eq:word2vec}. 

We present a trained 100-dimensional Word2Vec embedding on our publication dataset. For further analysis, we compute the average PUB-W embedding over all words.

\subsubsection{\textbf{PUB-G Embedding}}
We believe that the TFIDF may not capture the actual semantic similarity among words in the vector space. There has been numerous work on different types of embeddings proposed over the years, which capture better semantic meaning of textual information. With this motivation, we propose to use GloVe \cite{pennington2014glove} embedding trained on our publication corpus. In contrast to Word2Vec \cite{mikolov2013efficient} or TFIDF, GloVe does not only rely on the local context of the words. It captures global statistics by the means of the word-word co-occurrence matrix. Let us define the co-occurrence matrix as $X \in \mathcal{R}^{N\times N}$ where $N$ is the number of unique words in the dataset. $X_i,j$ is defined as the number of times the word $i$ has co-occurred with the word $j$. Let $X_i = \sum _k X_{i,k}$ be the number of times  any word appears in the context of word $i$. Furthermore, they define the $P_{i,j}$ as follows: 

\begin{equation}
    P_{i,j} = P(j|i) = \frac{X_{i,j}}{X_i}
\end{equation}

Let $w_i \in \mathcal{R}^d$ be the $d$ dimensional GloVe word embedding for the word $i$. They define a regression model to learn $P_{i,k}/P_{j,k}$. Here, learning the word embeddings depends on three words $i$, $j$ and $k$, where they define word $k$ to be the context word. Regression model is parameterized as follows: 

\begin{equation}
    F((w_i-w_j)^T \cdot \Tilde{w_k}) = \frac{P_{i,k}}{P_{j,k}}
\end{equation}

Here, for the context word $k$, separate embedding layer $\Tilde{w_k}$ is used. In our analysis, we use 100 as the dimensionality of the GloVe word embedding. 

For the abstract texts, we first compute the PUB-G embedding for each word in the abstract and then compute the average embedding over all the words.

\subsection{Clustering}
The resulting vectorized matrix obtained from feature vector space is then used to cluster each document of all four corpora, using K-Means Clustering~\cite{macqueen1967some}. To visualize these clusters in 2-dimensional space, we used t-distributed stochastic neighbor embedding (t-SNE)~\cite{maaten2008visualizing} visualization. Both of the approaches are discussed in detail further. 

\subsubsection{\textbf{K-means}}
~\cite{macqueen1967some} came up with one of the most straightforward unsupervised machine learning (ML) algorithms which are nowadays widely used for various real-world applications in order to recognize hidden patterns. It uses a simple approach of classifying the datapoints in a fixed number of clusters defined by $K$, each having a particular centroid, $c$ representing the center of the cluster. At first, the K-Means clustering algorithm initializes the random centroids followed by the recursive computations of the distance between each cluster point and its corresponding cluster's centroid until the centers of clusters get stabilized or the given number of iterations has been reached. The main idea is to minimize an error function known as a squared error which can be represented by the objective function \ref{eq:kmeans1} and the recalculation step of new centroid can be represented by \ref{eq:kmeans2}.

\begin{equation} \label{eq:kmeans1}
    J(V) = \sum_{i=1}^{c} \sum_{j=1}^{c_i} ( \norm{x_i - v_j})^2
\end{equation}

\begin{equation} \label{eq:kmeans2}
    v_i = (1/c_i) \sum_{j=1}^{c_i} x_i
\end{equation}

Here, $X = \{x_1, x_2, x_3,..., x_n\}$ is the set of datapoints whereas $V = \{v_1, v_2, v_3,..., v_c\}$ is the set of centroids of clusters. In our case, datapoint $x_i$ is a $T$-dimensional textual embedding for $i$\textsuperscript{th} abstract. The absolute difference between $x_i$ and $v_j$ shows the euclidean distance between the $i$\textsuperscript{th} abstract and $j$\textsuperscript{th} cluster, $c_i$ represents the number of datapoints in $i^{th}$ cluster, and $c$ depicts the total number of clusters used. A datapoint is assigned to a particular cluster based on the following function. 

\begin{equation}
    k_i = argmin_{j \in [1:K]}(||x_i - v_j||)^2 
\end{equation}

Here $k_i$ is the assigned cluster ID of $i$\textsuperscript{th} document.

We decided to use K=10 clusters across different publication datasets.
For our dataset, we used Elbow Method~\cite{yuan2019research} to first find the optimal number of clusters, leading to the earlier mentioned choice.

\subsubsection{\textbf{t-SNE}} 
\begin{figure*}[!th]
    \centering
    \begin{subfigure}[b]{0.485\textwidth}
        \centering
        \includegraphics[width=1\textwidth]{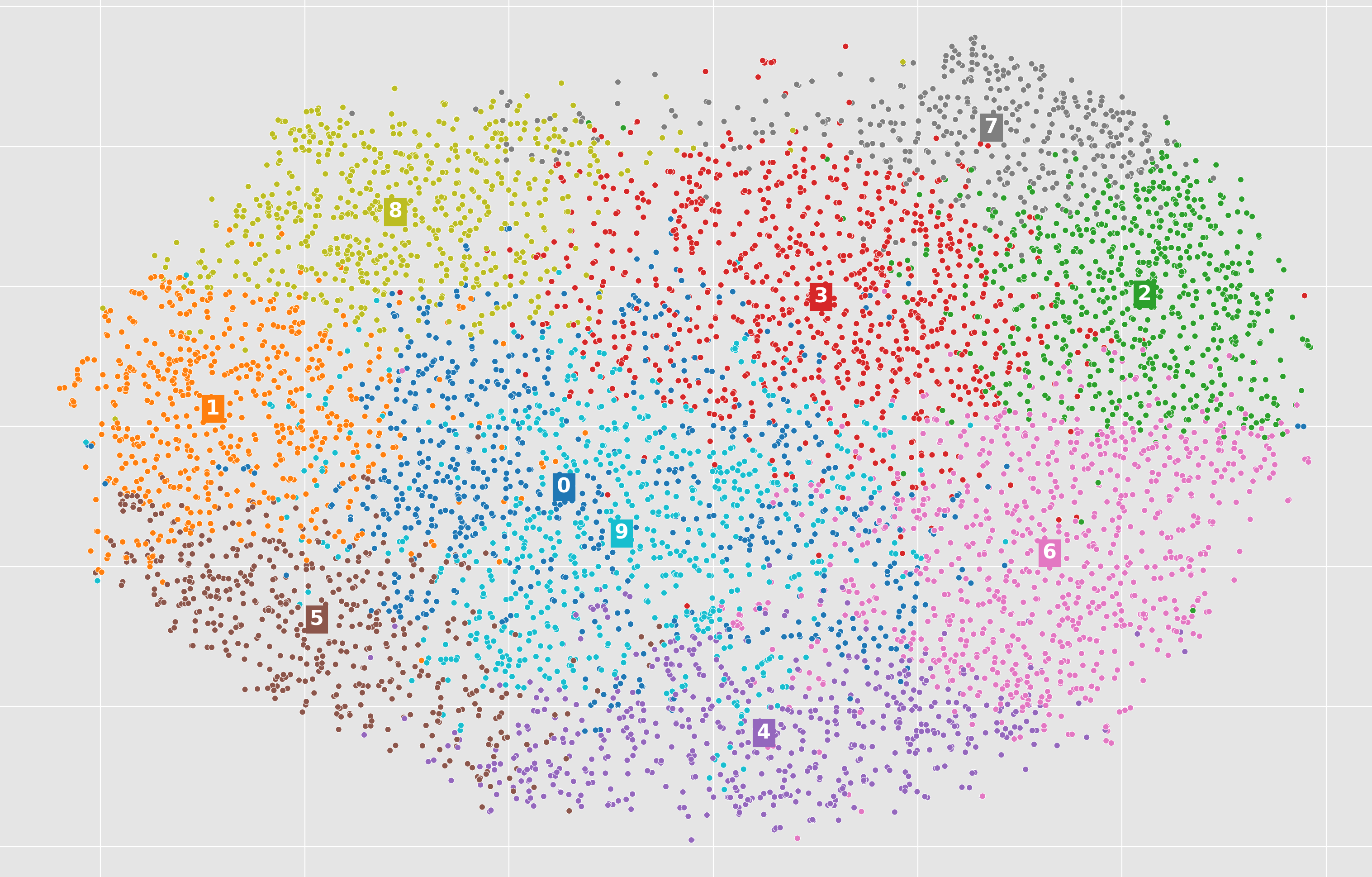}
        \caption[(a)]%
        {{\small}}    
    \end{subfigure}
    \hfill
    \begin{subfigure}[b]{0.485\textwidth}  
        \centering 
        \includegraphics[width=1\textwidth]{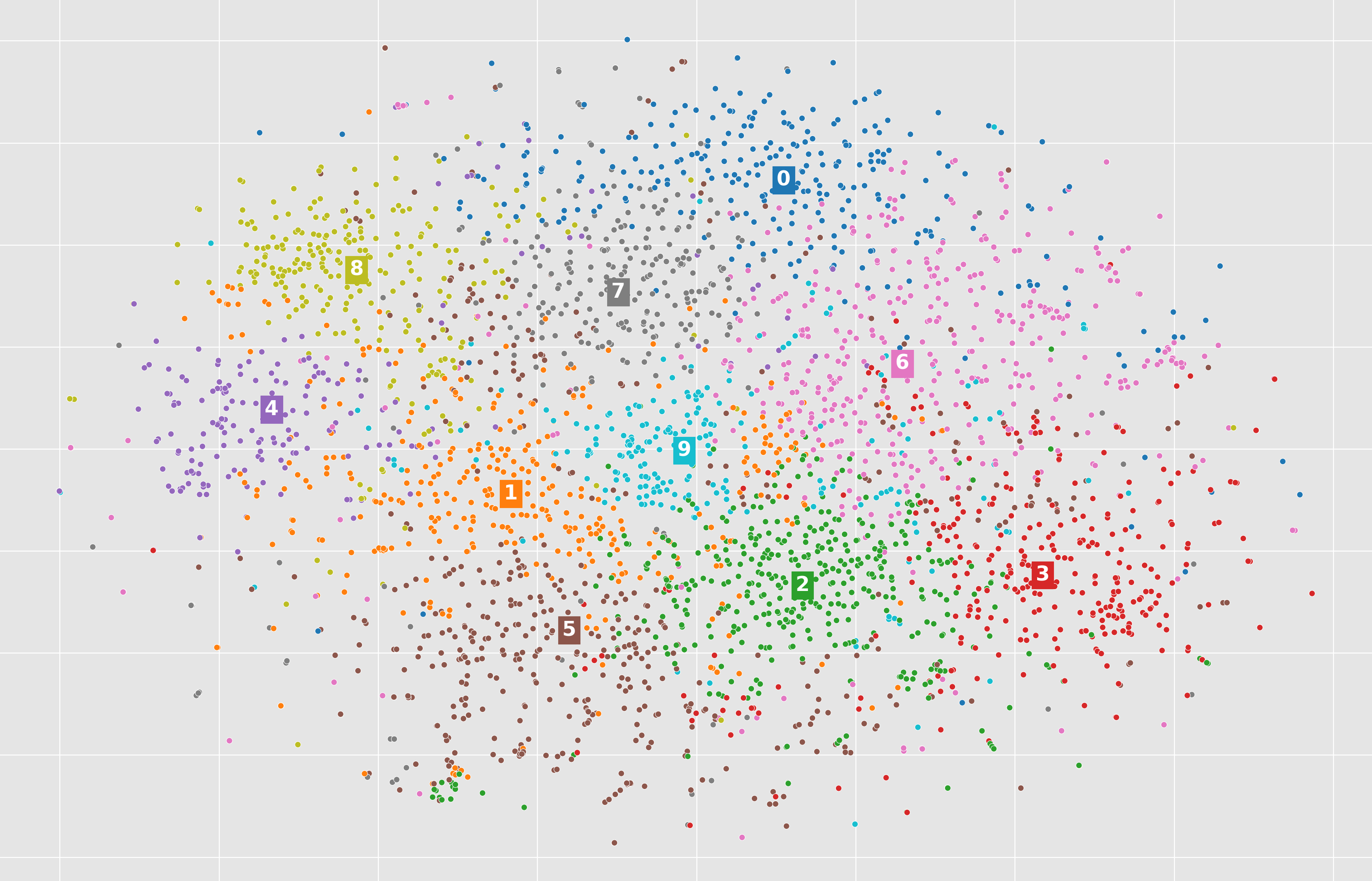}
        \caption[(b)]%
        {{\small}}    
    \end{subfigure}
        \begin{subfigure}[b]{0.485\textwidth}   
            \centering 
            \includegraphics[width=\textwidth]{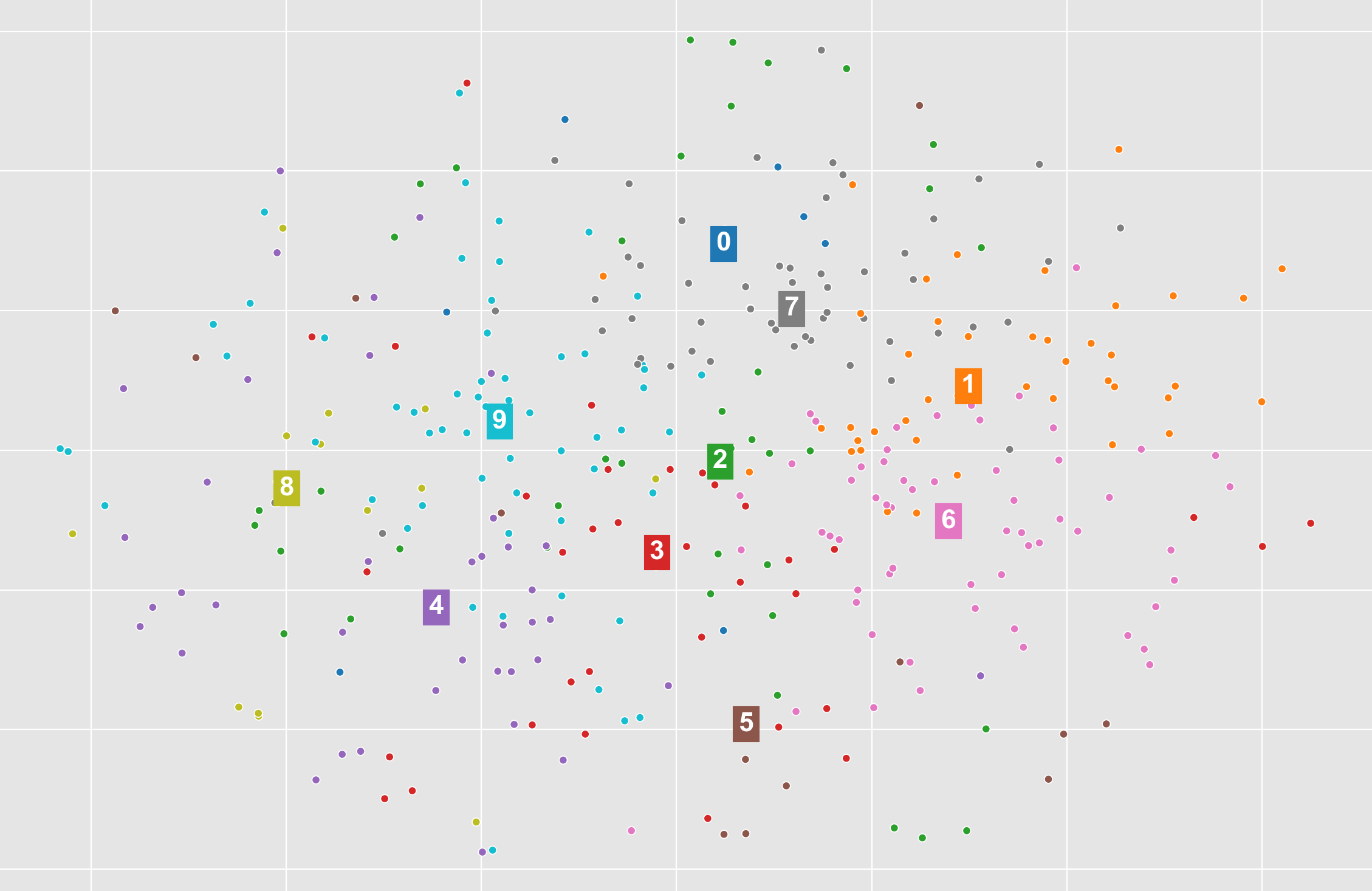}
            \caption[(c)]%
            {{\small}}    
        \end{subfigure}
        \hfill
        \begin{subfigure}[b]{0.485\textwidth}   
            \centering 
            \includegraphics[width=\textwidth]{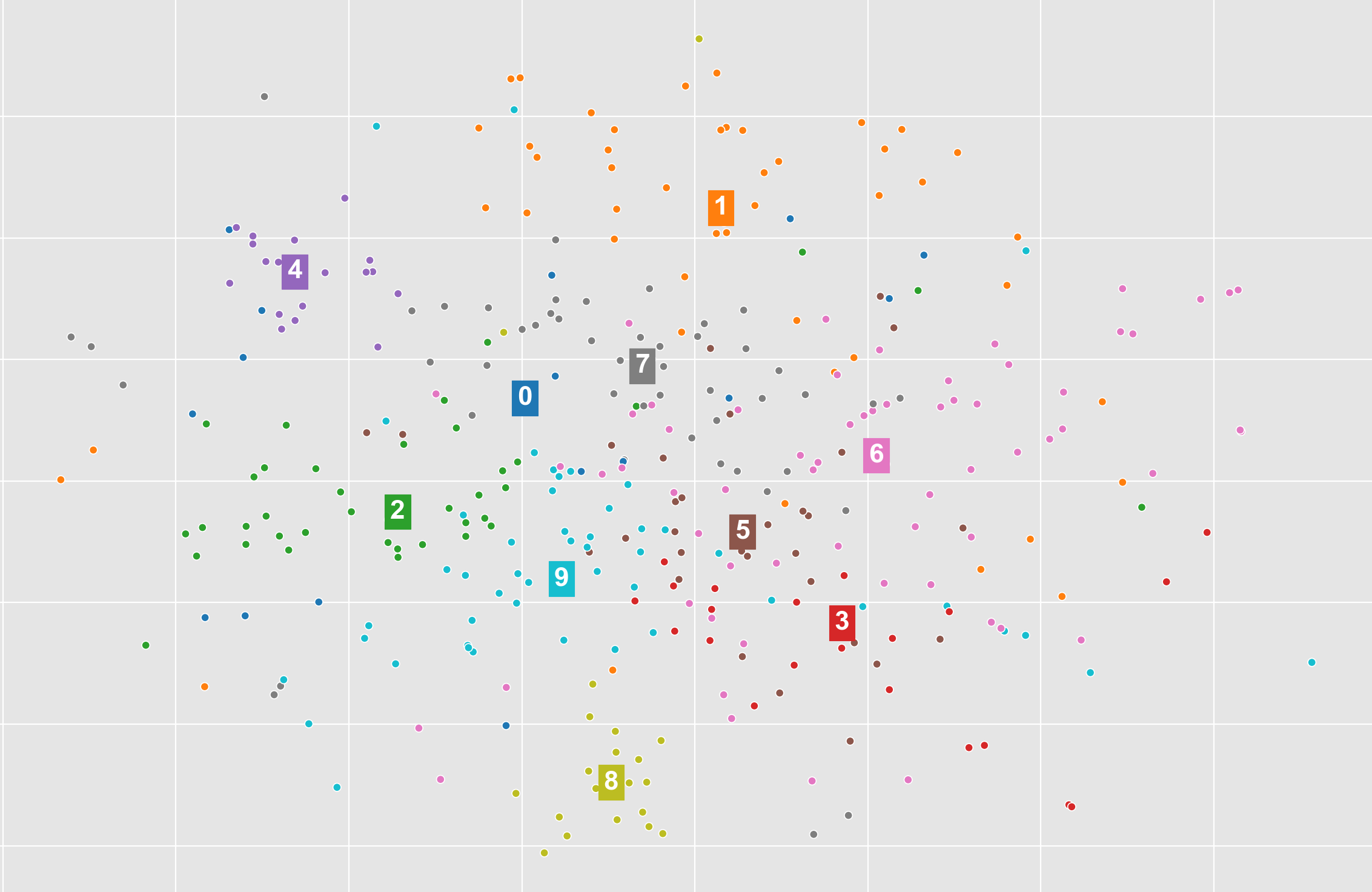}
            \caption[(d)]
            {{\small}}  
        \end{subfigure}
    \caption[ Clusters Visualization using t-SNE for (a) CHI; (b) ICED; (c) TOCHI; and (d) RIED based on PUB-G.]
    {\small Clusters Visualization using t-SNE for (a) CHI; (b) ICED; (c) TOCHI; and (d) RIED based on PUB-G.} 
   \label{fig:clusters}
\end{figure*}

Based on~\cite{hinton2002stochastic},~\cite{maaten2008visualizing} developed t-distributed stochastic neighbor embedding (t-SNE) that is extremely useful to visualize high-dimensional data in lower dimensions, specifically the two-dimensional plane. Hence, it is an unsupervised non-linear machine learning technique that is also used as a dimensionality reduction method. The t-SNE creates probability distribution by finding pairwise similarity between the neighboring datapoints. This pairwise similarity is decided based on the conditional probability density between two nearby points as it will be high for the nearby points rather than the far-distanced points. It contains two important parameters: perplexity and early exaggeration. Perplexity is the total number of nearest neighbors of the center point impacting the variance of Gaussian distribution whereas early exaggeration controls the space between the clusters. For our experiments, we kept the perplexity as 100 and early exaggeration as the default value i.e. 12.
Figure \ref{fig:clusters} shows the clusters visualizations graphs developed using t-SNE.
 
\subsection{Topic Modeling}

After clustering documents in groups, we discovered the topics to understand what each cluster is representing. We found out 10 topics for each cluster using the technique, LDA (Latent Dirichlet Allocation)~\cite{blei2003latent}. LDA is one of the widely used topic modeling approaches where it assumes that each document is a mixture of k different topic and each $k$\textsuperscript{th} topic has its inherent word probability distribution. Hence the objective of the LDA algorithm is to find these k topics and their word probability distribution. More concretely, let us assume that there are $D$ abstracts, $T$ words, $K$ topics, and $N$ words an abstract. The goal of LDA is to calculate the joint posterior probability as stated in the following equation:

\begin{equation}
    P(\theta _{1:D}, z_{1:N}, \beta _{1:K} | \mathcal{D}, \alpha _{1:M}, \eta _{1:K})
\end{equation}

Here $\theta$ is a distribution of topics, one for each document. $z$ is a distribution of topics, one for each word. $\beta$ is a distribution of words, one for each topic. $\alpha$ is the parameter vector for each document, $\eta$ is the parameter vector for each topic and $\mathcal{D}$ is the abstracts dataset.
\label{sec:lda}
    
\section{Dataset}
\label{dataset}

\subsection{Publication Venues}

We curate a dataset that represents the leading conferences and journals for the fields of Human-Computer Interaction and of Engineering Design, which are:
\begin{itemize}
    \item \textbf{CHI}. The ACM Conference on Human Factors in Computing (CHI) is one of the premier conferences for HCI research community. The conference deals with various topics like ubiquitous computing, visualization, usability, and user experience design in multiple tracks including long and short research papers, posters, workshops, and case studies.
    \item \textbf{TOCHI}. The ACM Transactions on Computer-Human Interaction (TOCHI) is a scientific journal covers a wide variety of topics dealing with software and hardware aspects of HCI field. This incorporates architectures, evaluation techniques, interactive interfaces, and user design processes.
    \item \textbf{ICED}. The International Conference on Engineering Design (ICED) is one of the main conferences for the engineering design research community. A wide range of topics are covered, creativity, AI and design, user-centered design, sustainable design, design-for-X, design theory, design methods and methodologies, human-behaviour in design, Industry 4.0, etc.
    \item \textbf{RIED}. Research in Engineering Design (RIED) publishes papers on design theory and methodology in all fields of engineering, focusing on mechanical, civil, architectural, and manufacturing engineering. Topics covered include functional representation, feature-based design, shape grammars, process design, redesign, product data base models, and empirical studies.
\end{itemize}

The selection of a journal and a conference series from two different fields, allows us to compare the usefulness of our methodology between fields, and between publication type. Table \ref{datasetDesp} shows a summary description of our dataset, which we elaborate more on later.

\begin{table}[!htbp]
\renewcommand*{\arraystretch}{1.25}
\centering
\caption{Summary of Dataset Statistic}
\label{datasetDesp}
\begin{tabular}{lll} 
\hline 
 &  & No. of\\
Venue & Year & Publications\\
\hline  
CHI & 2007--2020 & 6,365\\
TOCHI & 2007--2019 & 370\\
RIED & 1995--2020 & 366\\
ICED & 2003--2019 & 3,308\\
\hline
\end{tabular}
\end{table}

\subsection{Dataset Collection Procedure}

The Web of Science platform is used to fetch the data for CHI conference, TOCHI, and RIED journals. For CHI, publications from 2007 to 2020 were extracted, resulting in 6,365 data points. Similarly, the TOCHI dataset includes 370 data points, covering  2007 to 2019, while the RIED dataset contains 366 datapoints from 1995 to 2020 excluding 1996. The datasets include abstracts, titles, authors' names, and keywords.

To get the ICED Conference dataset, we extracted the publication information and papers from the ICED Design Society website\footnote{https://iced.designsociety.org/group/7/Proceedings+of+ICED} for the period 2003 to 2019. The total number of instances extracted is 3308. Due to the unavailability of some attributes on the website, the missing datapoints are manually extracted from the publications' PDFs. The dataset comprises (i) Title, (ii) Year, (iii) Editor, (iv) Author, (v) Series, (vi) Institution, (vii) Section, (viii) DOI number/ISBN, (ix) ISSN, (x) Abstract, and (xi) Keywords. The data is assembled as a delimited text file using a comma to separate values.

\section{Experimental Results and Discussion}
\label{results}

\subsection{Evaluation Metrics}
\subsubsection{\textbf{Topic Coherence}}

A set of statements or facts are said to be coherent if they support each other. As an example consider the statement `GloVe is a word embedding used for language modeling'. This statement is said to be coherent since the facts support each other. In literature, various topic coherence scores are used to quantify this semantic similarity. As explained in the section \ref{sec:lda}, we apply LDA topic modeling to each cluster to identify research topics. For a single topic, a coherence score measures the degree of similarity between high-scoring words in the topic. There exist two types of coherence metrics, 1) Intrinsic and 2) Extrinsic methods. Intrinsic methods do not use any external task for measuring semantic similarity. In comparison to that, extrinsic methods apply the discovered topics in an external task such as information retrieval. However, we believe is that applying the topics generated for a corpus abundant with domain-specific words into an external task is not well suited. Therefore, we use the intrinsic UMass coherence score \cite{mimno2011optimizing} for our evaluations. The following equation depicts the pairwise score function used to calculate the coherence score.

\begin{equation}
    score_{UMass}(w_i,w_j) = log \frac{D(w_i,w_j) + 1}{D(w_i)}
\end{equation}

Let us assume that, $w_i,...,w_n$ represents the top-n words for each topic. Here, $D(w_i)$ depicts the number of times the word $w_i$ appeared in the corpus and $D(w_i,w_j)$ depicts the number of times $w_i$ and $w_j$ appeared together in the corpus. Here, the $w_i$ is selected to be more common word than $w_j$.

Then the following score is averaged over all the topics and subsequently over all the clusters we generated.

\subsubsection{\textbf{Mutual Information Scores}}
We applied 3 different types of word embeddings, which generated distinct clusters. The mutual information (MI) measures the similarity between two labels of the same data as follows: 

\begin{equation}
    MI(U,V) = \sum _{i=1}^{|U|} {\sum _{j=1}^{|V|}{\frac{|U_i \cap V_j|}{N} log \frac{N|U_i \cap V_j|}{|U_i||V_j|}}}
\end{equation}

Here $U$ and $V$ are two different clusters, $|U|$ and $|V|$ are the number of cluster labels in cluster $U$ and $V$ respectively. $|U_i|$ is the number of samples in the cluster $U_i$ and $|V_j|$ is the number of samples in the cluster $V_j$

Furthermore, we also calculate two other MI related metrics called, 1) Normalized mutual information score (NMI) and 2) Adjusted mutual information score (AMI). NMI score scales the MI score to be between 0 and 1. AMI is another adjustment of MI score for chance. It is calculated as following: 

\begin{align}
    H(U) = - \sum _ {i=1}^{|U|} {\frac{U_i}{N} log(\frac{U_i}{N}}) \\
    H(V) = - \sum _ {i=1}^{|V|} {\frac{V_i}{N} log(\frac{V_i}{N}}) \\
    AMI(U,V) = \frac{MI(U,V) - E(MI(U,V))}{\frac{(H(U)+H(V))}{2} - E(MI(U,V))}
\end{align}

\begin{table}
\renewcommand*{\arraystretch}{1.25}
\centering
\caption{Average Coherence Score for all Publication Datasets}
\begin{tabular}{cccccc}
\hline
Publication Dataset & TF-IDF  & GloVe  & PUB-G & PUB-W \\ \hline
CHI                 & -1.7120 & -1.5677 & -1.5759  & -1.5200  \\ 
ICED                & -1.8409 & -1.5300 & -1.6537  & -1.5453 \\ 
TOCHI               & -6.2586 & -6.6287 & -5.5893  & -5.8207  \\ 
RIED                & -4.1561 & -4.0050 & -3.6394  & -3.7642 \\ \hline
\end{tabular}
\label{table:coherence}
\end{table}

\subsection{Comparing the Coherence score against various embeddings}
As explained in the subsection \ref{sec:lda}, we applied LDA topic modeling to identify 10 topics in each cluster separately. Then the average \emph{u\_mass} coherence score over the clusters for various text embeddings was computed. Table \ref{table:coherence} summarizes these values representing TF-IDF as the baseline, GloVe as the pre-trained, and PUB-G and PUB-W as the proposed embeddings. Based on these, we make the following key observations:

\begin{table}
\renewcommand*{\arraystretch}{1.25}
\centering
\caption{Mutual Information Scores for all Publication Datasets between TF-IDF and PUB-G Embeddings}
\begin{tabular}{cccc}
\hline
Dataset & M.I. Score & Adjusted M.I. Score & Normalized M.I. Score \\ \hline
CHI                 & 0.4608  &   0.2207              & 0.2231                \\ 
ICED                & 0.5670     & 0.2648             & 0.2692              \\ 
TOCHI               & 0.4450     & 0.1654            & 0.2117               \\ 
RIED                & 0.6882     & 0.2852              & 0.3246               \\ \hline
\end{tabular}
\label{tab:tfidf_glove}
\end{table}

\begin{table}
\renewcommand*{\arraystretch}{1.25}
\centering
\caption{Mutual Information Scores for all Publication Datasets between PUB-G and PUB-W Embeddings}
\begin{tabular}{cccc}
\hline
Dataset & M.I. Score & Adjusted M.I. Score & Normalized M.I. Score \\ \hline
CHI                 & 0.6274  &   0.2734             & 0.2754               \\ 
ICED                & 0.9640     & 0.4222            & 0.4254               \\
TOCHI               & 0.8318     & 0.3520          & 0.3872               \\
RIED                & 0.8371     & 0.3527             & 0.3877              \\  \hline
\end{tabular}
\label{tab:glove_word2vec}
\end{table}

\begin{table}
\renewcommand*{\arraystretch}{1.25}
\centering
\caption{Mutual Information Scores for all Publication Datasets between PUB-W and TF-IDF Embeddings}
\begin{tabular}{cccc}
\hline
Dataset & M.I. Score & Adjusted M.I. Score & Normalized M.I. Score \\ \hline
CHI                 & 0.6622  &   0.3185        & 0.3206               \\
ICED                & 0.5084     & 0.2368              & 0.2414               \\
TOCHI               & 0.4735     & 0.1697           & 0.2160               \\ 
RIED                & 0.7293    & 0.2948            & 0.3348 \\  \hline
\end{tabular}
\label{tab:word2vec_tfidf}
\end{table}

\begin{figure*} [!th]
    \centering
    \includegraphics[width=1\linewidth]{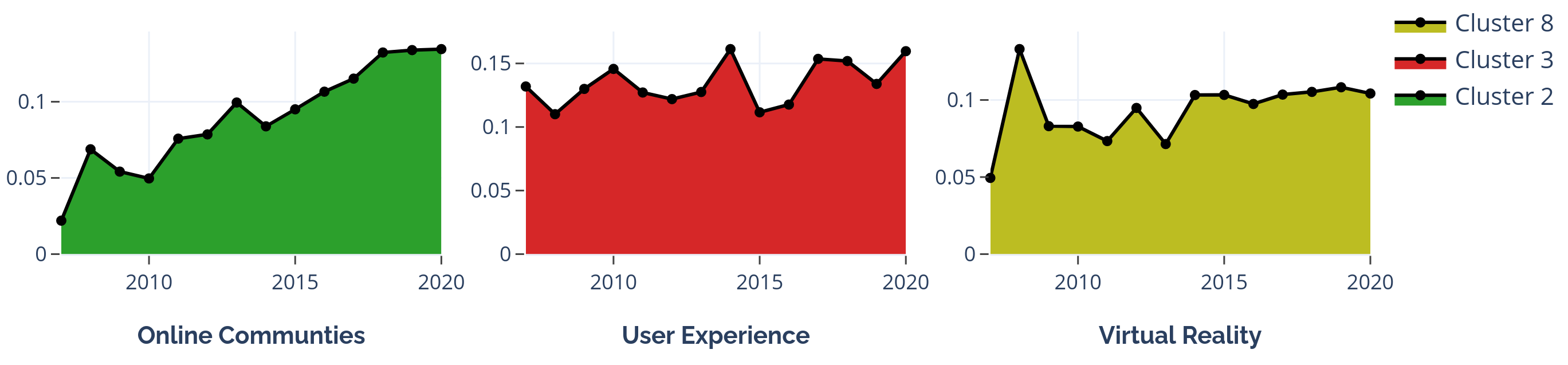}
    \caption{Trend analysis for CHI clusters}
    \label{fig:chi_selected_clusters}
\end{figure*}

\begin{enumerate}
    \item Our proposed PUB-G and PUB-W embeddings generates the best coherence score for all the publication datasets in comparison to all other embeddings. For CHI and ICED, PUB-W works the best whereas for TOCHI and RIED, PUB-G works well. PUB-G seems to generalize better in the cases of limited data points in comparison to PUB-W.
    \item Across all the datasets, TF-IDF performance is lower than both of the proposed, PUB-G and PUB-W embeddings. We believe that TF-IDF is not able to model the semantic meaning of the text.
    \item GloVe performs marginally better than PUB-G for ICED and CHI. However, GloVe's performance is still lower than our proposed embedding, PUB-W.
\end{enumerate}

Based on the key observations, we can notice that the already existing pre-trained textual embeddings are incapable of capturing the semantic meaning of textual data in a scientific domain.

\subsection{Comparing the MI scores between TFIDF, PUB-G, and PUB-W}

Different embeddings generated different clusters based on the inherent semantic similarity among words. In order to analyze the degree of similarity among these clusters, we compute the MI scores among these three embeddings. Tables \ref{tab:tfidf_glove}, \ref{tab:glove_word2vec}, and \ref{tab:word2vec_tfidf} shows the MI score between 1) TF-IDF and PUB-G, 2) PUB-G and PUB-W, and 3) PUB-W and TF-IDF. Based on these findings, we find that MI scores between PUB-W and PUB-G is higher in comparison to PUB-G and TF-IDF. This indicates that TF-IDF generates different clusters in comparison to the other two.

\subsection{Qualitative Analysis of clusters}

We demonstrate the qualitative analysis of the clusters detected by discussing a case study for the CHI publications. Upon applying the k-means clustering to all the documents and further applying topic modeling on the 10 clusters, we found 3 clusters out of 10 have shown some spikes, decrements, and steadiness from the year 2007 to 2020.

From the time-series graph in Figure \ref{fig:chi_selected_clusters}, it can be noticed that cluster 8 has shown a decrease which somewhat validates the fact as to when the first VR technologies came into existence, it had a sudden spike in interest in the research community but gradually decreased over the years and might decrease further due to COVID-19. On similar grounds, cluster 2 has shown a spike which can also be validated as data analysis of online communities, groups, forums have increased especially in terms of behavioral analysis, opinion mining, sentiment analysis, etc. 
Cluster 3, which mostly includes research on User Experience, appears to be having a steady but higher interest in the CHI community over the years. This may also be validated given that the theme of CHI research is focused on developing human-computer interactive technologies.

\section{Conclusion}
\label{conclusion}

In this paper, we present a framework to facilitate the scientific analyses of academic publications, which is important for monitoring the growth of a particular research field and identifying potential innovations. Our framework adopts and combines data collection, word embedding, topic modelling and temporal trend analysis. Many word embedding are trained on general text articles which may not be able to capture the features relevant to domain-specific texts found in scientific publications. To solve this problem, we curated a publication dataset consisting of two conferences and two journals from 1995 to 2020 in two research disciplines. Using this dataset, we propose two scientific publication embedding, i.e., PUB-G and PUB-W, which are capable of learning semantic meanings of general as well as domain-specific words in various research fields. Experimental results show that our PUB-G and PUB-W embeddings out-perform other baseline embeddings based on topic coherence.

\section{Acknowledgements}
This research is funded in part by the Singapore University of Technology and Design under grant SRG-ISTD-2018-140. The authors thank the anonymous reviewers for their useful comments.

\balance
\bibliographystyle{IEEEtran}
\bibliography{paperEmbed}

\end{document}